\documentclass{ecai} 

\usepackage{latexsym}
\usepackage{amssymb}
\usepackage{amsmath}
\usepackage{amsthm}
\usepackage{booktabs}
\usepackage{enumitem}
\usepackage{graphicx}
\usepackage{color}

\newcommand{\BibTeX}{B\kern-.05em{\sc i\kern-.025em b}\kern-.08em\TeX}

\begin{document}


\begin{frontmatter}


\paperid{123} 


\title{Playing Hex and Counter Wargames using Reinforcement Learning and Recurrent Neural Networks}


\author[A]{\fnms{Guilherme}~\snm{Palma}\thanks{Corresponding Author. Email: guilherme.palma@tecnico.ulisboa.pt}}
\author[A,C]{\fnms{Pedro A.}~\snm{Santos}\orcid{0000-0002-1369-0085}\thanks{Corresponding Author. Email: pedro.santos@tecnico.ulisboa.pt }}
\author[B,C,D]{\fnms{João}~\snm{Dias}\orcid{0000-0002-1653-1821}\thanks{Corresponding Author. Email: joao.dias@gaips.inesc-id.pt}} 

\address[A]{Instituto Superior Técnico, University of Lisbon}
\address[B]{Faculty of Science and Technology, University of Algarve}
\address[C]{INESC-ID}
\address[D]{CISCA}


\begin{abstract}
Hex and Counter Wargames are adversarial two-player simulations of real military conflicts requiring complex strategic decision-making. Unlike classical board games, these games feature intricate terrain/unit interactions, unit stacking, large maps of varying sizes, and simultaneous move and combat decisions involving hundreds of units. This paper introduces a novel system designed to address the strategic complexity of Hex and Counter Wargames by integrating cutting-edge advancements in Recurrent Neural Networks with AlphaZero, a reliable modern Reinforcement Learning algorithm. The system utilizes a new Neural Network architecture developed from existing research, incorporating innovative state and action representations tailored to these specific game environments. With minimal training, our solution has shown promising results in typical scenarios, demonstrating the ability to generalize across different terrain and tactical situations. Additionally, we explore the system's potential to scale to larger map sizes. The developed system is openly accessible, facilitating continued research and exploration within this challenging domain.
\end{abstract}

\end{frontmatter}


\section{Introduction}

The pursuit of perfect or super-human proficiency in board games has long been a benchmark for assessing the capabilities of contemporary knowledge and technologies in artificial intelligence. Historically, significant milestones have been achieved in games such as Chess, Othello, and Backgammon, where AI models have either matched or surpassed the best human players~\cite{othello,gnu_backgammon,alphaZero}. Additionally, considerable advancements have also been noted in strategic negotiation games like Diplomacy~\cite{diplomacy_paper}. Despite these successes, there are still numerous even more complex games that still elude good AI-based solutions.
This paper explores the application of advanced AI techniques to a category of board games known as Hex and Counter Wargames. These are adversarial two-player simulations of real military conflicts that often involve complex scenarios requiring strategic decision-making. Unlike classical board games such as Chess and Go, these games feature intricate terrain/unit interactions, unit stacking, large maps of varying sizes, and simultaneous move and combat decisions involving hundreds of units.

Our methodology is based on Alpha Zero \cite{alphaZero}, but with a substantially different Neural Network architecture, adapted to much lower computing requirements and the specific needs of these types of games. The main contributions of our work are: 
\begin{itemize}
    \item \textbf{Network architecture} - A dual head fully convolutional Recurrent Neural Network, adapted to this type of games, that can be used with boards of varying dimensions and needs less training resources.

    \item \textbf{State and action representations} - Novel representations which can be utilized on many Hex and Counter Wargames, and that can be extended to accommodate specific game needs.
    
    \item \textbf{Generalization results} - Significant results on the capacity of this type of architecture to generalize to different board and unit configurations serve as important groundwork for future research.
    
    \item \textbf{Game environment and AlphaZero implementation} - We developed an environment for Hex and Counter Wargames, and a asynchronous AlphaZero implementation, based on the original paper, with extra features adapted to these specific games. This framework is freely available for use by other researchers\footnote{The code is available at the Github repository: \url{https://github.com/guilherme439/NuZero}}.
\end{itemize}


\subsection{Problem Setting}

Hex and Counter Wargames are adversarial boardgames played on hexagonally tiled maps that use different cardboard markers and counters to represent the several units and resources available on the map. These games are usually, but not limited to, simulations of historic military conflicts, commonly including large maps that represent multiple regions or countries, and rulebooks with many dozens or sometimes hundreds of pages (e.g.: \cite{TSWW,OCS_rules4.2,Stalingrad42}). 

All the training and testing throughout this study was conducted under a sub-set of rules inspired by the Standard Combat Series ruleset \cite{SCS_rules1.8}, that are common to a large amount of Hex and Counter Wargames and which we will now describe. 

The games are played on game boards that can be configured to have any desired dimensions. Each tile on the game board has a type of terrain with 3 attributes: attack modifier, defence modifier and movement cost. Additionally, certain tiles on the map yield victory points.

Play lasts for a predetermined number of turns. After these turns, the player who controls the highest percentage of victory point tiles emerges victorious. 

Throughout the game, players place units on the map that they receive as reinforcements. Units are characterized by their three features: attack strength, defence strength and movement allowance. Groups of units can stack in the same tile up to a predefined limit. Beyond that, there are no stacking group restrictions, meaning that all units can stack with each other.

Combat can be conducted between units on adjacent hexes. When a group of units attacks another, the total attack and defence strength of each group of units is calculated to determine the combat's result. For the attackers, each unit multiplies its attack strength by the terrain's attack modifier and adds it to the group's total. The same thing happens in the group of defending units, using the defence strength and modifiers. The group that obtains the lower combined strength loses their strongest unit. If a draw happens, both groups lose a unit. 

Each player's turn consists of two distinct phases: the movement and combat phases. Both the movement and combat phases comprise two sub-phases, resulting in a total of four sub-phases throughout each player's turn. During the movement phase, players engage in the reinforcement and movement sub-phases, wherein they strategically place reinforcements and manoeuvre their units, respectively. Meanwhile, the combat phase entails sub-phases for: target selection (where the player chooses the tile to attack) and attacker selection (in which adjacent units are chosen to enter combat). 

Each turn is characterized by both players engaging in both phases, with each player completing their turn before the other can proceed. The only exception to this rule happens at the beginning of the game, before the first turn, where both players place their initial reinforcements on the map, without going through any other phases.

\section{Related Work}


The solution we developed has strong foundations on previous research. The system is mostly based on the work done by \textit{Deepmind} with AlphaZero \cite{alphaZero}, and the research provided by \textit{Deep Thinking systems} with the papers in \cite{easy_to_hard} and \cite{deepthinking_2}. In this section, we provide a more in-depth look at those systems and finish with a brief overview of previous work done in similar games.

\subsection{AlphaZero}

Deepmind developed a series of algorithms to play complex board games. Each of these algorithms iterated upon the research obtained from the previous solutions. AlphaZero is one of those algorithms and mostly bases itself on the previous systems: AlphaGo \cite{alphago} and AlphaGo Zero \cite{alphago_zero}. These systems demonstrated great success, as they achieved super-human play in their respective domain. However, while the two first systems were designed solely for the game of Go, AlphaZero is a general reinforcement learning algorithm that can be applied to a large set of different games.

AlphaZero's algorithm makes use of a neural network with two heads: a policy head and a value head. Given a neural network policy prediction $\pi$ and value prediction $v$, the algorithm utilizes a modified Monte Carlo Tree Search (MCTS) to obtain both an improved policy and new value estimates. These two new quantities are then used to update the neural network, which can, once again, be used to provide new value and policy predictions ($v$ and $\pi$), repeating the cycle, to obtain an increasingly better network.

To achieve this AlphaZero utilizes two independent phases: network training and self-play data generation. These two phases can happen asynchronously, and they only exchange information by: sending the most recent version of the network from training to self-play through a shared network storage, and by sending the completed games from self-play to training via a replay buffer.

\subsection{Deep Thinking}

A group of researchers from the University of Maryland developed \textit{Deep Thinking systems} to study methods of achieving logic extrapolation using Recurrent Neural Networks. The team published two papers \cite{deepthinking_2, easy_to_hard}, that study three distinct problems: prefix sum computation, mazes, and chess.

For each of these problems, a fully convolutional neural network was trained employing a recurrent module that can be repeatedly used during inference for arbitrary numbers of iterations. The training is conducted on small/simple problems, utilizing a relatively low number of iterations and, afterwards, the network is tested on larger or more complex problems using larger numbers of recurrent module iterations in inference.

This research demonstrated the capacity to obtain models that, not only extrapolate to larger problems but also increase their performance with higher numbers of recurrent iterations.

In the second paper \cite{deepthinking_2}, the researchers addressed some of the limitations introduced in the first publication \cite{easy_to_hard} and looked to achieve even better performance by creating an improved network architecture and a new training methodology, both of which we now describe.

\begin{figure}[h]
\begin{center}
\includegraphics[width=0.49\textwidth]{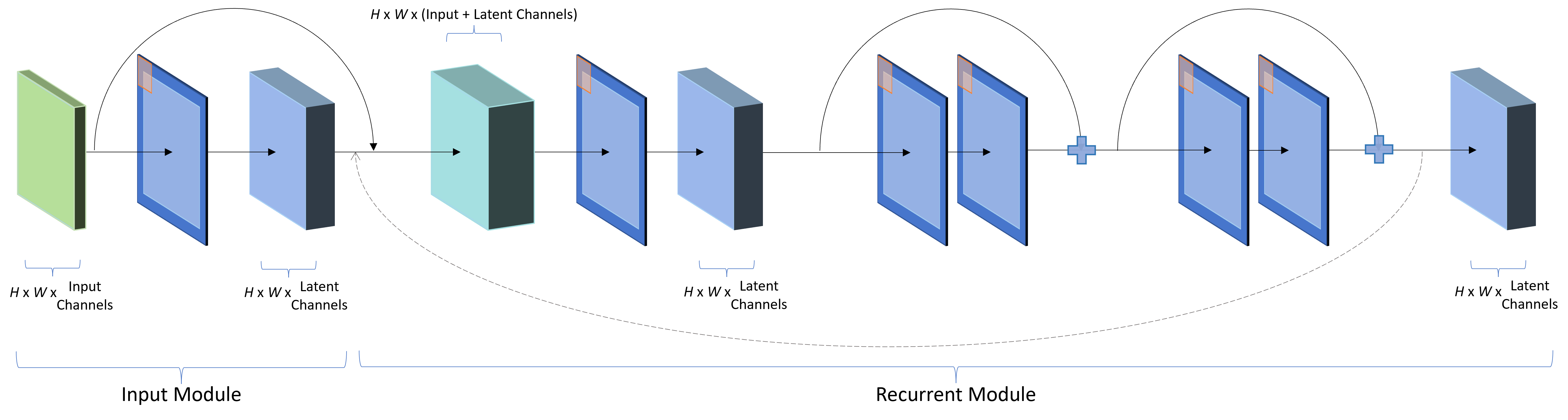}
\caption{Input and recurrent modules of the \textit{Recall} architecture.}\label{input_and_recurrent_modules}
\end{center}
\end{figure}

\textit{Recall} was the name given to the new architecture (Figure \ref{input_and_recurrent_modules}), and it is characterized by a new connection that concatenates the input to the beginning of the recurrent module, followed by a convolutional layer to compress the representation back to the original number of channels.

The new training methodology modified loss calculation by introducing ``progressive loss``. To calculate the progressive loss we choose two numbers at random $n$ and $k$, such that $ (n + k) < T$, where $T$ is the maximum number of iterations we wish to perform. We start by passing the input through the input module and through the recurrent module for $n$ iterations, discarding all the gradients. Then we pass the outcome of this operation through the recurrent module for $k$ more iterations and finally through the output module. In the end, we obtain the progressive loss by calculating the loss function between this last output and the target. This progressive loss is then combined with the regular loss obtained from the output after the maximum number of iterations, using a parameter $\alpha$:

\begin{equation}\label{eq:progressive_loss}
L = (1 - \alpha) \times L_{maxIters} + \alpha \times L_{progressive}
\end{equation}

\subsection{Previous Work}

The Wargame landscape remains very diversified, despite attempts at unification and formalization \cite{formalizing_wargames}, with environments ranging from realistic air combat simulation for pilot training \cite{multi_agent_rl_coordination_air_combat, adaptive_behavior_modeling_air_combat} to simple family board games like Risk \cite{Risk_info}.


Several properties of these games make them particularly challenging for AI approaches \cite{ai_and_wargaming}, leading a lot of research to focus on human-crafted knowledge as a way to address this obstacles, either by directly enforcing optimal actions in certain board states \cite{PK_DQN_paper}, or by simplifying in-game features using hand-crafted functions \cite{MADM_wargames}. More recently, however, \cite{MADDPG_wargames} has demonstrated promising results, utilizing new methods and relying on human knowledge just to accelerate early training using Supervised Learning.

Other research have focused on the usage of Reinforcement Learning, such as the paper \cite{wargame_combat}, which provided important contributions to the study of unit combat in wargames. This research was conducted on a 10x10 featureless map and demonstrated the impact of different combat models and policy optimization algorithms on the training of Reinforcement Learning agents.


On the other hand, in 2019, Glennn Moy and Slava Shekh published  research~\cite{Alpha_wargames} that proposed the use of AlphaZero in Wargaming. This paper showed that AlphaZero was capable of reaching optimal play in a simple 10x10 environment with two units and a single terrain type, but demonstrated that human-crafted heuristics were still needed in order to accelerate training.

There have also been several master thesis released by students of the Naval Post Graduate School \cite{cnn_combat_nps,naval_movement_rl_nps,feudal_rl_nps}, which address Wargames of a very similar nature with techniques such as Feudal Reinforcement Learning, Deep Q-Networks and AlphaZero.

Beyond Wargames, there have been other games where similar Reinforcement Learning techniques have been utilized, such as \textit{FreeCiv} \cite{free_civ_paper} and \textit{Terra Mistica}. The research in \cite{AlphaTM} applied AlphaZero to the game of \textit{Terra Mistica} and gives important insights into possible action and state space representations, since the game shares multiple characteristics with Wargames such as the adversarial nature, hexagon-based boards and multi-phased turns.

\section{Methods}

\subsection{Training Strategy}

The main concept of our solution is the combination of the techniques discussed in \cite{deepthinking_2, easy_to_hard} with the learning algorithm of AlphaZero~\cite{alphaZero}.

\begin{figure}[h]
\begin{center}
\includegraphics[width=0.478\textwidth]{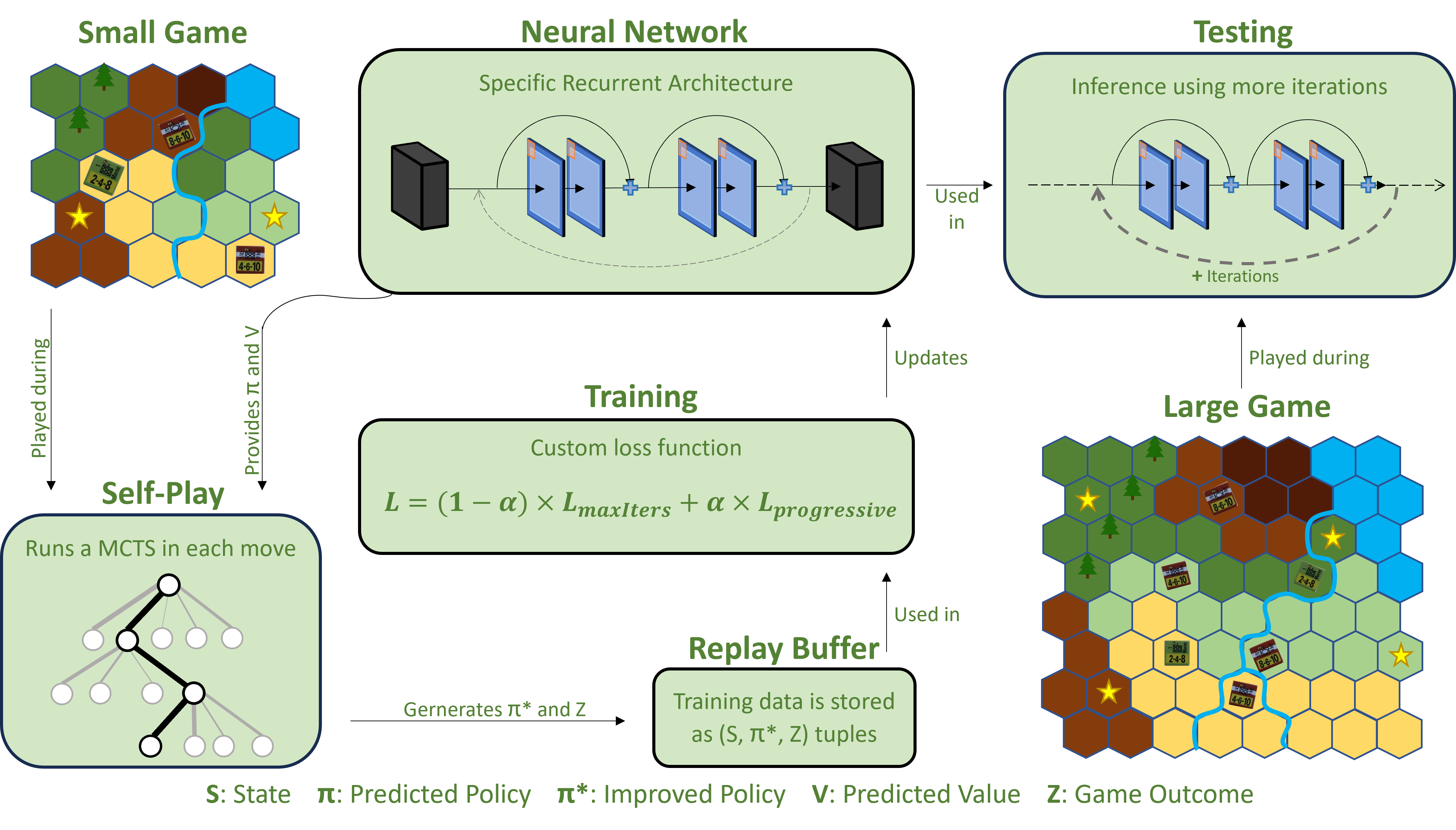}
\caption{Diagram of the designed solution}\label{training}
\end{center}
\end{figure}

The first key aspect is the utilization of the recurrent neural network architecture described in \cite{deepthinking_2}. This architecture, allows us to control how many times information goes through the recurrent module, before being passed to the output module (the number of recurrent iterations), while simultaneously being fully convolutional, which enables the use of different input sizes/resolutions. 

Having this in mind, we use AlphaZero's algorithm to first train the network on a small game map, using a low number of recurrent iterations. This not only decreases training time but also allows the algorithm to train in a simple environment, with a smaller search space, where it can learn strategies with greater ease.


When training ends, we test that trained network on a larger map, but we use a larger number of recurrent iterations. This is the same methodology used in \cite{deepthinking_2, easy_to_hard} and it should allow the network to extrapolate what it learned on the small map, to the larger one, simply by increasing the number of recurrent iterations. This process was designed with the intent of obtaining networks capable of playing on larger scenarios using the limited computational resources we had available.

\subsection{Game State Representation and Network Input}

The game state representation is a very important part of our solution since we wish to have a representation capable of encompassing a broad set of Hex and Counter Wargames, without requiring re-training of the network. 

Since Hex and Counter Wargames are oftentimes highly asymmetric and will regularly have different win conditions for each player, we have decided to represent board positions statically for the two players instead of always using the perspective of the current player as in systems like AlphaGo.

The game state was represented using a feature stack with the same resolution as the board ($Height \times Width$) with different channels representing different properties of the game.

The following sections, explain how each of the game characteristics was represented. We start by outlining the features that are necessary for both players (win conditions, unit positioning and  reinforcements), and then take a closer look at the remaining properties that are still required to fully represent the game. 

\subsubsection{Win conditions}

Players win through the control of victory point tiles. For each player, this is represented using $1$ channel, that one-hot encodes the victory point locations on the board.

\subsubsection{Unit positioning}

The units present on the board need to be represented taking in consideration their different attributes, statuses and also their position and stacking. 

There is a group of planes for each of the 3 unit statuses: ready to move, already moved and already performed an attack. Within each status group, units are represented using 3 channels, one for each of their attributes (attack strength, defence strength and movement points). In each of these channels the respective unit attribute is marked at each unit's position. However, since there are $S$ stacking levels, $S$ groups of 3 planes are required instead, per status. This leads to  $ 3 \times 3 \times S = 9S $ planes to represent units, for each player.

\subsubsection{Reinforcements}

Depicting the reinforcement schedule for the entire game would be too extensive. For this reason, we illustrate instead, the next $R$ reinforcements to come into play. We represent these units by the order in which they will come into play, using two sets of planes: the first portrays the units themselves, while the second represents how long they take to arrive. In the first set each unit is represented using 3 planes (one for each unit attribute), in each of those planes the respective unit attribute is marked at its possible arrival locations, while the remaining locations are marked with zeros. On the second set, there is an equal number of planes for each unit but they instead contain: the total number of turns minus the number of turns until the unit arrives\footnote{Using the number of turns until the unit arrives directly, would lead to the need of representing an ``infinite`` number of turns at the end of the reinforcement schedule.}. This representation results in a total of $( 3 \times R ) \times 2 = 6R$ channels, per player, to represent incoming reinforcements.

\subsubsection{Other features and final representation}

Considering terrain as the first player-independent aspect, we chose to represent it using 3 planes: one for each terrain attribute (attack modifier, defence modifier and movement cost).

Due to its complexity, the combat phase also requires some special attention, since there is the need to distinguish between different states that occur during the combat itself. First, we need to represent the state where a player has decided to attack a tile. This is accomplished by using $1$ channel where the position being attacked will be marked. We also need $S$ more channels marking the position of each of the units that the attacker has chosen to use, considering their stacking level. This leads to $S+1$ channels for state representation during combat.

Beyond combat and terrain, there are still some other features that are part of the game state: the current player, current turn and current sub-phase. To represent the sub-phases we add 4 more planes (one for each sub-phase). During each sub-phase the respective plane is filled with ones, while the remainder stays empty. For the current player, we add a plane filled with either 1 or -1, representing players 1 or 2 respectively. Finally, for the current turn we use one more plane filled with the current turn divided by the total number of turns, which can be interpreted as the percentage of the game played so far. 

All the player-independent features just described amount to a total of $3 + S + 7 = S + 10$ input channels. The features required by both players, previously described require $ 2\times(1 + 9S + 6R)$ more input channels. Putting everything together, we reach the final expression for the number of input channels of: $ 19S + 12(R+1)$



\subsection{Action Space Representation}

Actions were spatially represented using three dimensions, as we reasoned that such a representation might make the learning process for the neural network more direct, due to the adoption of convolutional layers in the neural network architecture. The three dimensional representation is achieved using several planes with the same resolution as the board ($Height \times Width$) stacked together.

We will start by considering the actions required for the movement sub-phase. Since pathing is such an important part of strategic planning, and movement plans are often complex, we have decided that movement decisions should be taken one tile at a time. Considering that we are moving over hexagonal tiles, each having $6$ adjacent neighbours, there will be $6$ groups of $S$ planes used to represent the movement. Each group represents one of the possible directions of movement, and each plane within that group will be used to select the position and stacking level of the unit to be moved. This representation takes $6 \times S$ planes. 

The fighting phase is perhaps the one that requires a more complex representation. Due to the different sub-phases, combats are executed using a sequence of several actions. First, to start an engagement the attacking player must choose the grid position he wants to attack. This is done using a single plane where the position is picked.  Next, the player selects one by one, each of the adjacent units he wants to use in the attack. For this purpose, $S$ planes are used so that the attacking player can select units according to both their position and stacking level. Finally, when the attacking player is ready, he must once again select the position he wishes to attack on yet another plane, in order to confirm the attack being made. All these sub-actions leave us with $1 + S + 1 = S + 2$ more planes.

\begin{figure}[h]
\begin{center}
\includegraphics[width=0.35\textwidth]{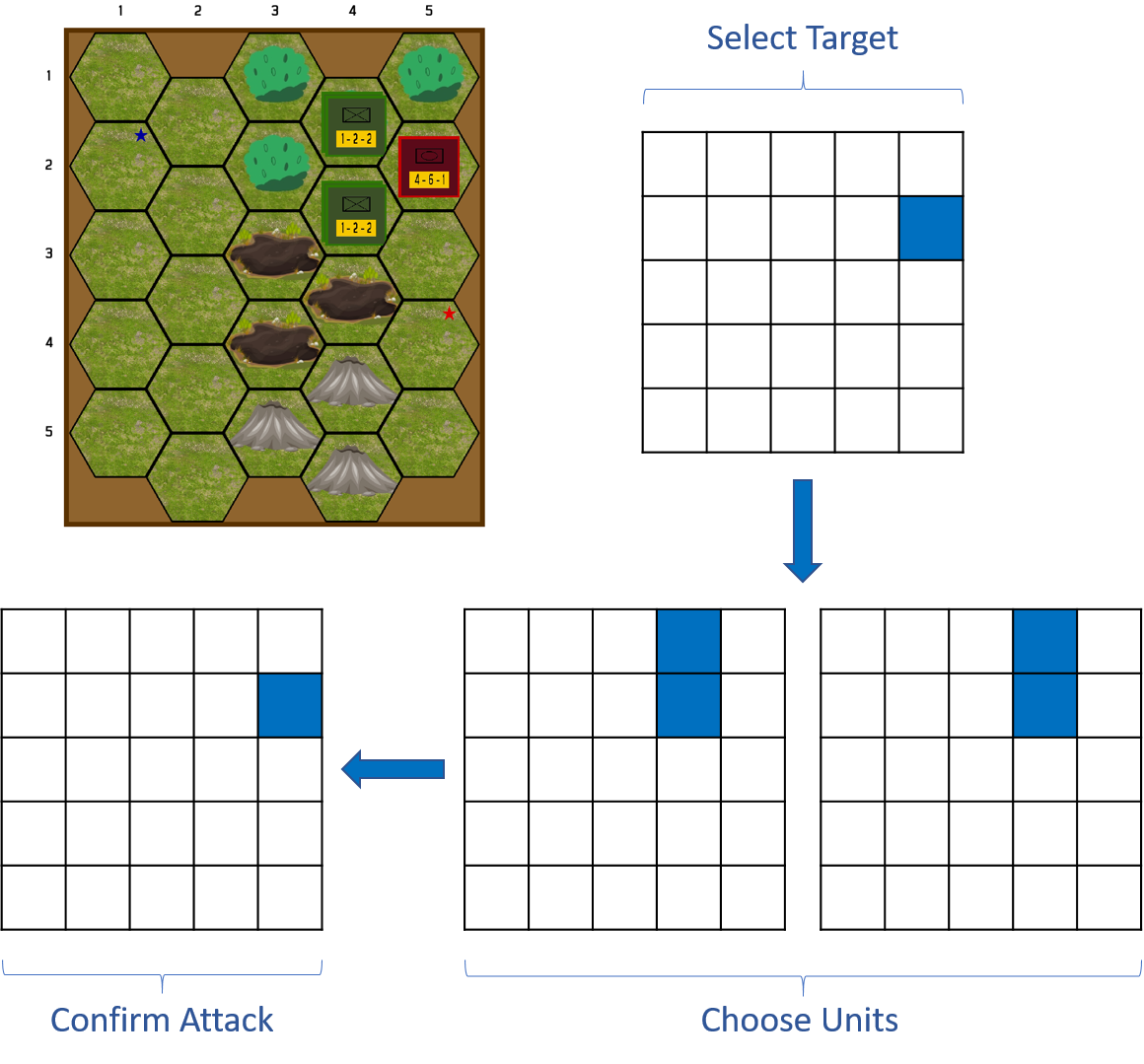}
\caption{Representation of an attack where all player one units (green) decide to target the enemy armour (red). Note that each action (blue square) would be taken sequentially. }\label{attack_representation}
\end{center}
\end{figure}

We are still missing the actions needed for reinforcements placement at the beginning of the movement phase. We can achieve this using a single plane where the player just needs to select a position to place the next incoming unit. 

Lastly, we need to add $S$ planes for the action of not moving a unit, and another $S$ more planes for the action of not attacking with a unit, for the cases where the player wishes to declare that we will not move or attack with certain troops, in a given turn. $S$ planes are used for each of these actions since there can be multiple units in the same location and we need to distinguish between stacking levels.

If we combine all the previously described planes we get an action space representation of size $ (9S + 3) \times Height \times Width$.

\subsection{Neural Network Architecture} \label{subsection:architecture}

The network architecture that was used is based on the \textit{Recall} architecture from the paper \cite{deepthinking_2}. Similar to the networks described there, our network can be divided into three sections: input module, recurrent module and output module. For the input and recurrent modules we maintained the original architecture (Figure~\ref{input_and_recurrent_modules}), while for the output module we use two output heads (one for the value and another for the policy), instead of just one.

\begin{figure}[h]
\begin{center}
\includegraphics[width=0.48\textwidth]{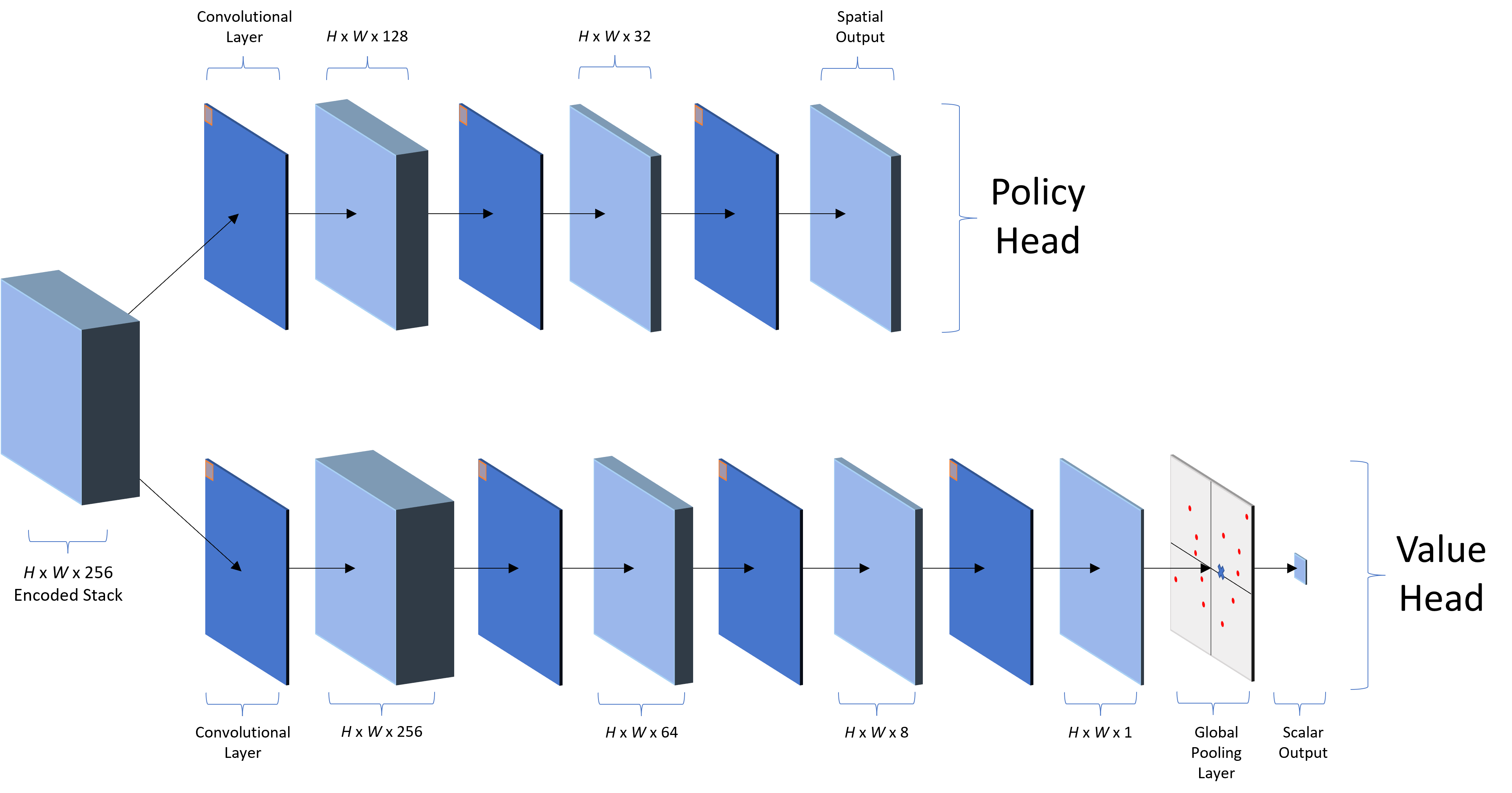}
\vspace{-2pt}
\caption{Output module architecture}\label{output module}
\end{center}
\end{figure}

We maintain a latent space with 256 channels throughout the network by using 256 filters on all the convolutional layers until we reach the output module.

Both the policy and value heads of the output module we designed can be seen in Figure~\ref{output module}. The value head consists of four convolutional layers that progressively reduce the number of filters, followed by an average pooling layer to reduce the output to a single scalar value. This value is then passed through a tanh activation function to give the final output.

The policy head contains three convolutional layers, each reducing the number of filters from the previous layer, with the final layer having as many filters as the number of action space channels.

Note that performing regular convolutions over hexagonally sampled data would deform the output of the operations. For this reason, since we are dealing with hexagonal tiles, all the convolutional layers in this network are hexagonal convolution layers, implemented using the Hexagdly library \cite{HexagDLy}.


\section{Results}

In order to test the system, three representative scenarios were used, an asymmetrical scenario, a symmetrical scenario (with 2 variants), and a randomized scenario. Each of these scenarios will be described in more detail in the following subsections.

We present first the results obtained by the neural networks during training and in tests done on the original maps, and afterwards, we analyse the network performance when extrapolating to larger game boards. All the networks presented used 6 recurrent iterations during training and a progressive loss $\alpha$ of 0.01.

The training for all the scenarios was done on a $5\times5$ game board, including two unit types and 4 possible terrain types. The unit types used were infantry and armour. On an even playing field, armour is stronger and more mobile than infantry, however, if two infantry units attack the armour together, or if they make use of terrain strategically, they can destroy it. On the other hand, the terrain types used where: plains, mountains, swamps, and forests. The terrains were designed so that plains are neutral, mountains and forest provide either a defensive or offensive advantage respectively, while swaps are poor defensively and hard to traverse.

All the training and testing was performed using only {CPU} since, at the time of training, the system was not optimized to efficiently use {GPU} and took little advantage of the extra hardware. 

To access the performance of our solution in all the different scenarios, using various opponents, four agents were created:
\begin{itemize}
    \item \textit{Policy} agent - An agent that selects the action which was given the highest probability by the network being used.
    \item \textit{MCTS} agent - This agent runs AlphaZero's MCTS before each action and selects the action with the highest visit count.
    \item \textit{Random} agent - The agent selects actions randomly amongst the possible actions in the current board state.
    \item \textit{Goal Rush} agent - This agent places units randomly and tries to move each unit towards the closest victory point. If an enemy stands in the path of a unit to its victory point, it attacks the enemy with all available units.
\end{itemize}

In the scenarios where the \textit{Goal Rush} agent was utilized 250 test games were played to determine each of the win rate values.
On the other hand, in the extrapolation graphs presented at the end of the this section, each data point is the average of 3 runs with 100 test games in each.


\subsection{Training Results}

\subsubsection{Asymmetric Scenario}

\begin{figure}[h]
\begin{center}
\includegraphics[width=0.175\textwidth]{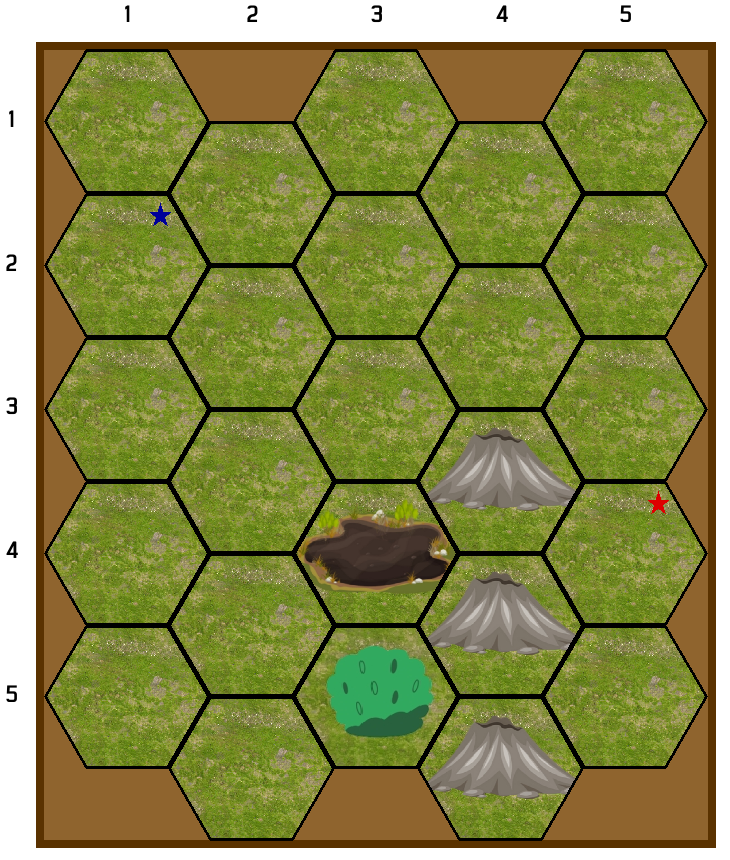}
\hspace{20pt}
\includegraphics[width=0.175\textwidth]{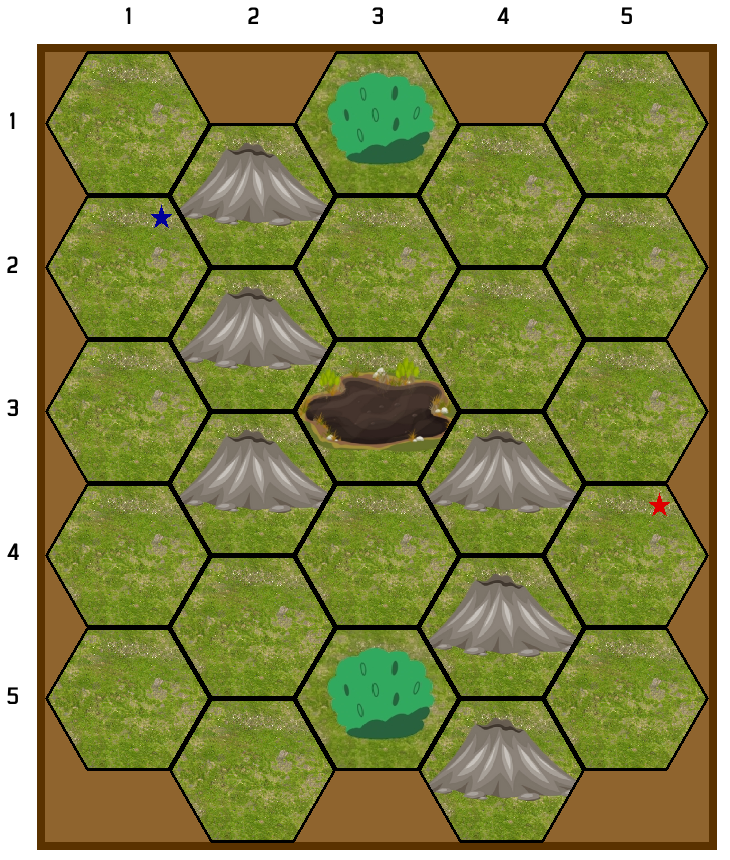}
\caption{Visualization of the terrain for the asymmetric scenario (left) and symmetric scenario (right).}\label{map_visualization}
\end{center}
\end{figure}

In the asymmetrical scenario, player one possesses two infantry units while player two only controls one. The player in disadvantage has a line of mountains in front of its victory point which provides a defensive advantage and allows it to draw if the enemy makes a mistake (Figure \ref{map_visualization}). The objective of this scenario is to understand if the system can learn in unfair situations.

\begin{figure}[h]
\begin{center}
\includegraphics[width=0.23\textwidth]{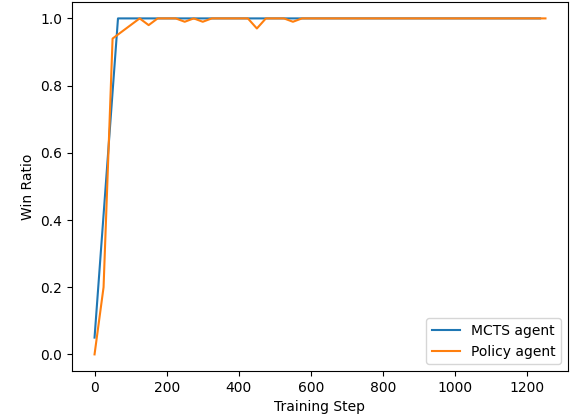}
\includegraphics[width=0.23\textwidth]{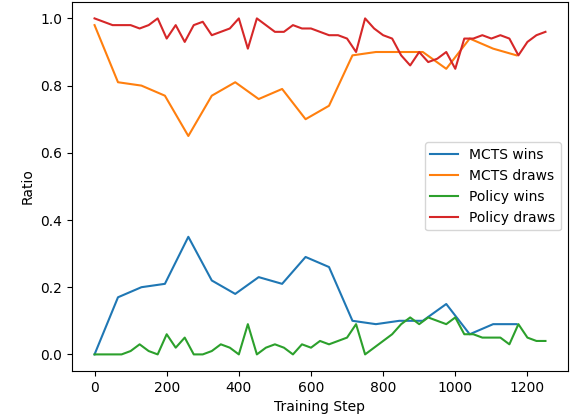}
\caption{Win rates obtained by player 1 (left) and player 2 (right), against the \textit{Random} agent during training, on the asymmetric scenario. For player two the figure also displays the ratio of draws.}\label{unbalanced_training}
\vspace{-10pt} 
\end{center}
\end{figure}

In this scenario, the network seemed to learn with greater ease the strategy for player one than for player two. The results (Figure \ref{unbalanced_training}) show that, during training, the agent easily learns player one's strategy, having a 100\% win rate against the \textit{Random} agent since very early on in training, while for player two it reached a maximum of 35\% of wins. The network was updated 16 times per training step using mini-batches of 256 positions. Attempts to further train the network were made, however the results did not improve.

Since player two is at a disadvantage in most situations, the optimal action is to stand in a good defensive position and wait for the opponent to make a mistake. For this reason, player two rarely observes states where it has an advantage and can push for a win, potentially causing it to have more difficulty in learning its strategy when compared to player one.

At the end of the training, we measured the performance of the network against the \textit{Goal Rush} agent, using each of the players, both with the \textit{MCTS} and \textit{Policy} agents. Once again, player one achieves a win ratio of 100\%, since the strategy employed by the \textit{Goal Rush} agent of rushing for the objective is very ineffective when at a disadvantage and the AI agent easily exploits it. 

\begin{table}[h]
\begin{center}
\caption{Win rate results from testing against the \textit{Goal Rush} agent in the asymmetrical scenario.}
\vspace{20pt}
\includegraphics[width=0.33\textwidth]{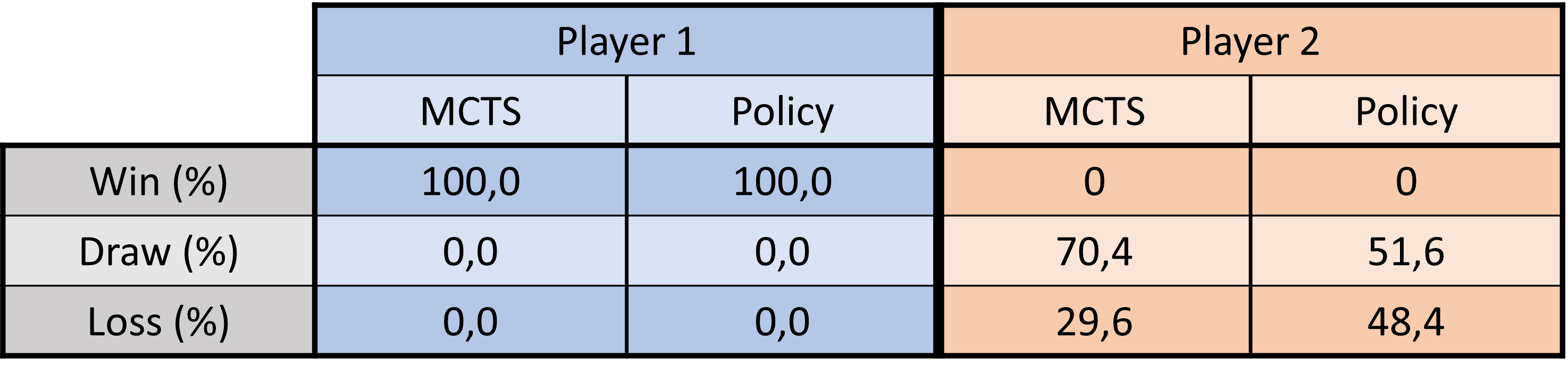}
\label{unbalanced_goal_table}
\end{center}
\end{table}

Meanwhile, with player two, when versing the attack by the \textit{Goal Rush} agent, the network manages to draw approximately 70\% and 51\% of the games with the \textit{MCTS} and \textit{Policy} agents respectively. For comparison, the \textit{Random} agent loses approximately 96\% of games and draws the remaining 4\% when playing against the \textit{Goal Rush} agent, demonstrating that despite not reaching perfect results, this network still obtained a very large improvement over random play.

\subsubsection{Symmetric Scenario}

For this map configuration, both players have two infantry units and the board is symmetric (Figure~\ref{map_visualization}). The objective of this game scenario is to check whether the agent can understand how to play the game in a very even scenario, where most games end in draws, but still win when playing against weaker opponents.

The network achieved very high win rates both using the \textit{Policy} and \textit{MCTS} agents. Training was conducted using training steps with 8 mini-batches of 512 positions.

\begin{figure}[h]
\begin{center}
\includegraphics[width=0.23\textwidth]{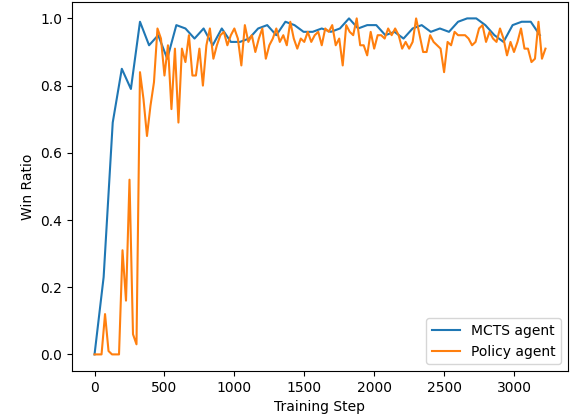}
\includegraphics[width=0.23\textwidth]{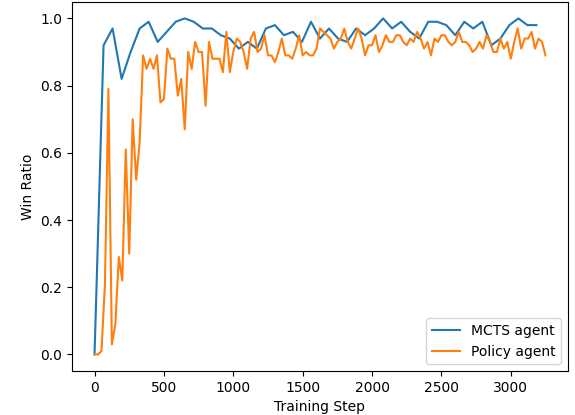}
\caption{Win rates obtained by player 1 (left) and player 2 (right), against the \textit{Random} agent during training, on the symmetric scenario.}\label{symmetric_training}
\end{center}
\end{figure}

Just like in the previous scenario the \textit{Goal Rush} agent was also used as a benchmark and the results for these tests can be seen in Table~\ref{symmetric_goal_table_both}. It should be noted that both \textit{MCTS} and \textit{Policy} agents managed to win the vast majority of the games and that neither of them lost any games against this agent.

\subsubsection{Symmetric Scenario - Curriculum Learning}

To test the effectiveness of curriculum learning, a new scenario was created based on the symmetric setting. This new scenario used the same map, however, each of the players received two extra reinforcements, one at turn 3 and another at turn 5. This effectively doubles the amount of units that come into play from 4 to 8. 

\begin{figure}[h]
\begin{center}
\includegraphics[width=0.23\textwidth]{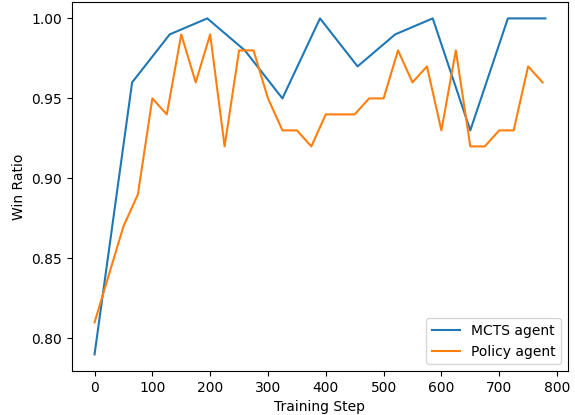}
\includegraphics[width=0.23\textwidth]{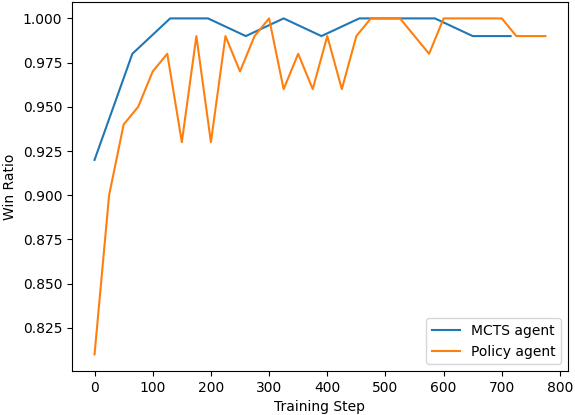}
\caption{Win rates obtained by player 1 (left) and player 2 (right), against the \textit{Random} agent during training, on the curriculum learning extension of the symmetric scenario. }\label{symmetric_cl_training}
\end{center}
\end{figure}

The training configuration used for this scenario did 32 network updates per training step, with a mini-batch size of 128.
We started the training using the latest network from the original symmetric scenario and after a short number of training steps, it could already achieve very high win rates against the \textit{Random} agent. These results demonstrate that curriculum learning could be used as an effective tool to increase the complexity, for game scenarios where there is already a well-trained network. 

Finally, the network was tested against the \textit{Goal Rush} agent (Table \ref{symmetric_goal_table_both}), where it obtained even better results than in the original scenario. The more units a player has compared to the opponent, the easier it is to overpower enemy units while taking fewer casualties. One of the reasons for the better performance in this scenario is the fact that, as the AI agent coordinates its units better, it can grow larger and larger leads and defeat the opponent with greater ease. The results also show that the \textit{Policy} agent has a higher win rate when playing as player two. After analysing some of the games between the two agents, we found that this happens because the AI tends to have a more defensive strategy when playing with player two, which works much better against the attacks from the \textit{Goal Rush} agent.

\begin{table}[h]
\begin{center}
\caption{Win rate results from testing against the \textit{Goal Rush} agent in both variations of the symmetric scenario.}
\vspace{20pt}
\includegraphics[width=0.46\textwidth]{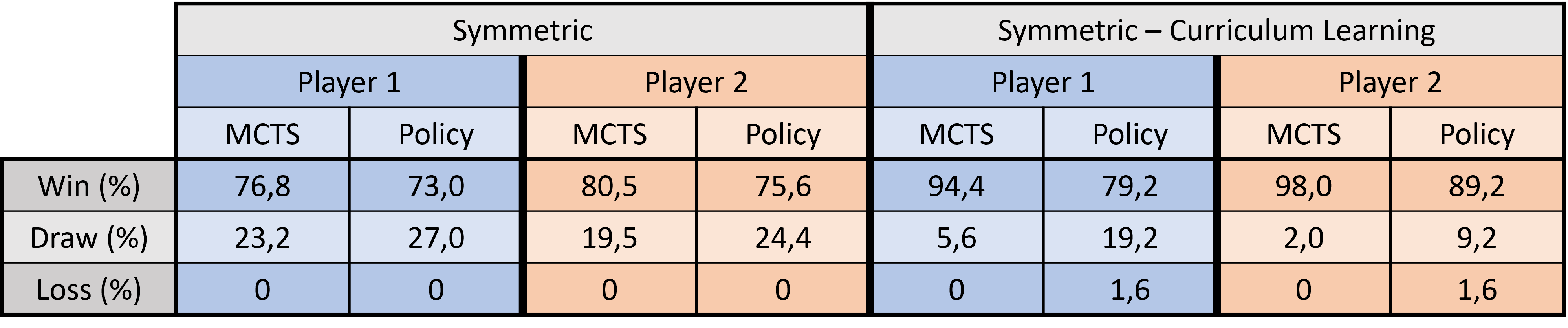}
\label{symmetric_goal_table_both}
\end{center}
\end{table}
\vspace{-5pt}

\subsubsection{Randomized Scenario}

For the final and most complex scenario, both players start with two infantry units and they receive an extra armour unit at the beginning of the second turn. The map is fully randomly generated using a distribution over the four terrain types\footnote{ The exact distribution used, had probabilities of 0.1, 0.15, 0.65, and 0.1 for swamp, mountains, plains, and forest, respectively.}. The objective of this scenario is to understand if the network can learn in a variety of different maps while managing different unit types.

The neural network used for this scenario had the longest training process, reaching a total of 3640 training steps with 32 weight updates within, each using a mini-batch of 256 positions.

\begin{figure}[h]
\begin{center}
\includegraphics[width=0.23\textwidth]{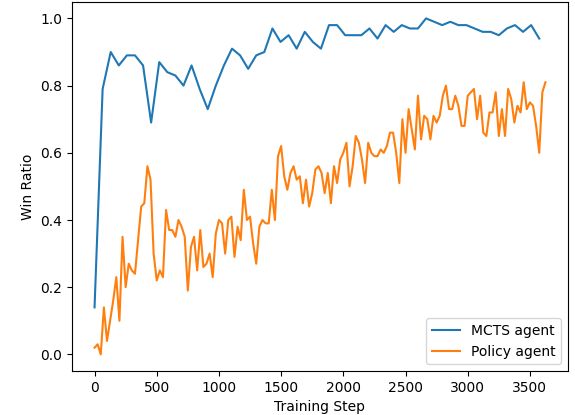}
\includegraphics[width=0.23\textwidth]{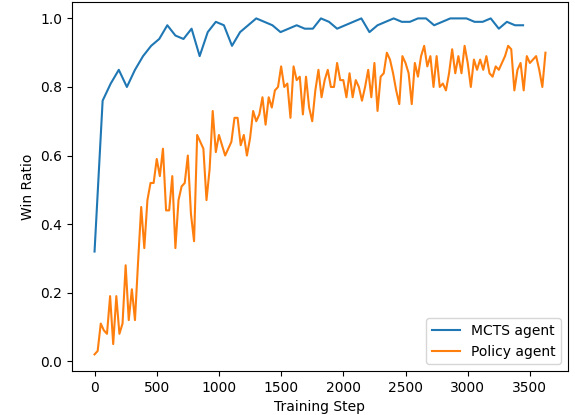}
\caption{Win rates obtained by player 1 (left) and player 2 (right), against the \textit{Random} agent during training, on the randomized scenario.}\label{randomized_training}
\end{center}
\end{figure}

During training, the network managed to achieve approximately 100\% win rate using the \textit{MCTS} agent and roughly 80\% with the \textit{Policy} agent for both players, versus the \textit{Random} agent. The performance against the \textit{Random} agent, especially for the \textit{Policy} agent, was slightly better on player two. This might have happened because, since the game always ends in player two's turn, it directly "sees" terminal states when using the MCTS during self-play, while player one always has to rely on the neural network predictions to evaluate the states.

\begin{table}[h]
\begin{center}
\caption{Win rate results from testing against the \textit{Goal Rush} agent in the randomized scenario.}
\vspace{20pt}
\includegraphics[width=0.33\textwidth]{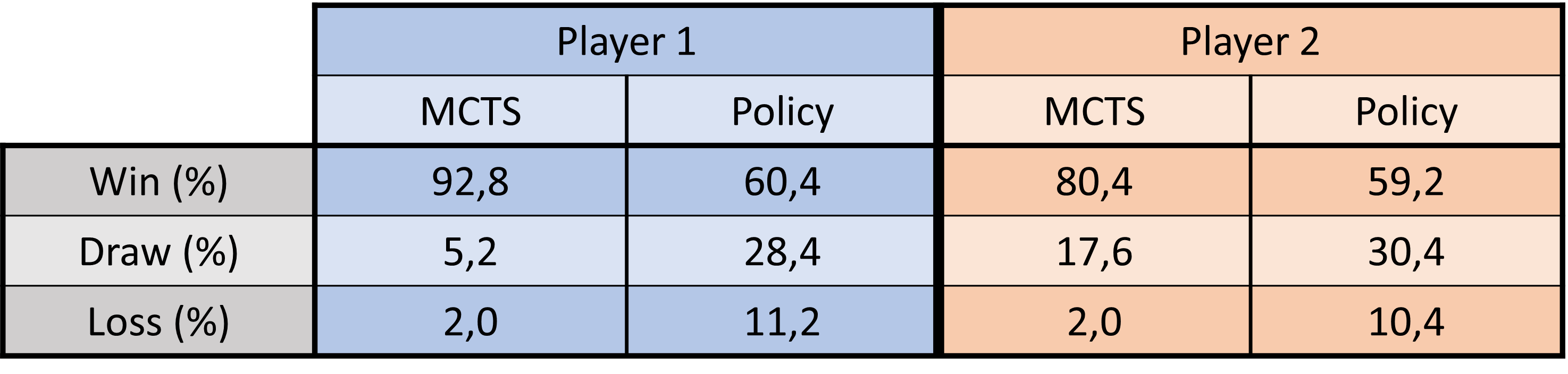}
\label{randomized_goal_table}
\end{center}
\end{table}

Similarly to the other scenarios, the \textit{Goal Rush} agent was also tested (Table \ref{randomized_goal_table}). The \textit{MCTS} agents managed to get above 92\% and 80\% win rates for players one and two, respectively. While the \textit{Policy} agents had a drop in performance to about 60\% win rate for both players. The difference found in \textit{MCTS} agent's win ratios from 92,8\% for player one to 80\% in player two, can be explained by the fact that, when playing on a random map, using intelligent agents, the capacity to act first is a big advantage. Since player one always plays first, it can dictate what kind of engagements happen, while player two is forced to simply react. On top of that, this player also receives its reinforcements first, which provides a very strong offensive advantage and can make certain positions indefensible for the enemy. When analysing games where the \textit{Goal Rush} agent played as player one, we noticed that in many situations, it would use the extra reinforcement to overpower the enemy defence, and the best that the \textit{MCTS} agent could do was counter-attack the enemy victory point and make a draw. Situations like these can explain the difference found in win rates, as we can see that the draw percentage increases in an equal amount to the win rate reduction. The randomized board and victory points from this last scenario, force the neural network to understand game strategy and not just ``memorize`` the best action in each state, leading to the display of more advanced strategic abilities. 


\subsection{Extrapolation to larger map sizes} \label{section:extrapolation}

After training the system on the 5x5 map, several tests were done on larger maps to test the system's extrapolation capacity. We started by testing on a map made solely of plains tiles, using one unit and a victory point on a random location. We trained multiple networks using different hyper-parameters, such as the number of recurrent module iterations and the $\alpha$ value for the progressive loss. Afterwards, the networks were tested on maps of increasing size (between 5x5 and 12x12). To test the effectiveness of the recurrent architecture itself, a residual neural network was also trained to serve as a baseline. This network had 12 residual blocks, which is the same effective depth as a RNN using 6 recurrent iterations (the lowest that we tested). Effective depth is defined as the depth that the recurrent network would have if we unrolled the recurrent iterations.

\begin{figure}[h]
\begin{center}
\hspace{10pt}
\includegraphics[width=0.40\textwidth]{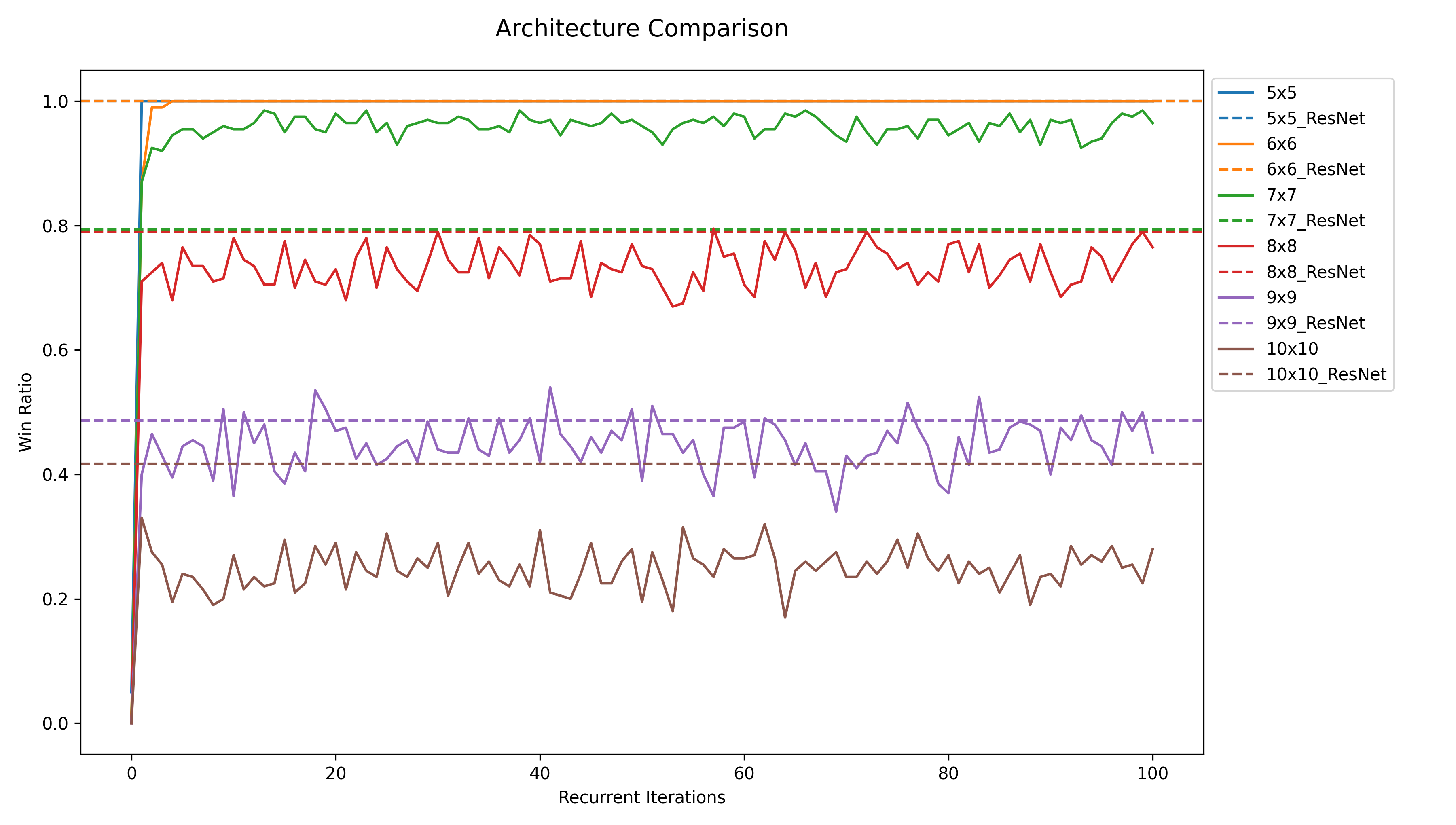}
\caption{Comparison between the win rates of the best performing recurrent network and the baseline residual network,  using varying numbers of iterations and different map sizes.}\label{rec_vs_res}
\end{center}
\end{figure}

Figure~\ref{rec_vs_res} presents the results obtained by the best performing recurrent network when compared to the baseline residual network. The best-performing recurrent network utilized 30 recurrent iterations (resulting in a much higher effective depth), and employed a progressive loss $\alpha$ value of 1, meaning that it only used progressive loss during training.

The new architecture obtained inferior results to the baseline residual network. We can also observe that the network's performance significantly decreases with the map size and, unlike what was expected, larger numbers of iterations do not seem to have any impact on the network performance.

We believe the issues found may be related to the nature of the problem and the way the network attempts to solve larger problems with more iterations. When considering a maze-like problem, if we are trying to solve a large maze, possessing information about what paths are or not viable in a smaller version of that maze, gives us direct and valuable information on how to solve the larger problem. However, for other more complex problems such as chess or the games we studied, the translation of information from simpler problems to more complex ones might not be so direct.

Another possible explanation for the poor extrapolation capacity might be the total amount of training that the neural networks undertook. The networks trained for our extrapolation tests ran for 1100 training steps and the maximum number of games held in the replay buffer was 6000, due to memory constraints, which pales in comparison to AlphaZero's 700,000 training steps and 1 million games' replay buffer. It could be the case that additional training would be required for the networks to extrapolate to larger map sizes.

\section{Conclusions and Future Work}

We proposed a system to play Hex and Counter Wargames. Our approach used AlphaZero's training algorithm and a recurrent network architecture based on the previous research \cite{deepthinking_2, easy_to_hard}. We obtained good results on several scenarios using small amounts of training and limited resources. We demonstrated, that the current methods (including curriculum learning) can generalize for variable terrain configurations and number of units in more complex scenarios, but are not enough to extrapolate well to larger map sizes. 

Regarding the neural networks used in this work, we presented a value head architecture designed for fully convolutional networks and the task of attributing values to different board states. Our architecture proved capable of learning in the scenarios we trained it on, even with variable terrain. Still, we believe that the incorporation of different layer architectures such as depth-wise or separable convolutions, is worth investigating in future research, as they could be a better fit to the problem's requirements. 

On the other hand, there has been a lot of other research done on the concept of adapting computation time to problem complexity \cite{MEMO, PonderNet, adaptive_early_exit, reinforce_rnn, ACT}. Due to large map sizes, usually displayed by Wargames, we consider that the integration of techniques such as PonderNet in similar systems could be a very promising avenue for future work.


\begin{ack}
The authors dedicate this paper to the memory of Dean Essig (1961-2024)
\end{ack}



\bibliography{mybibfile}

\end{document}